%% file: 128-main.tex
\documentclass[runningheads]{llncs}
\usepackage{makeidx}
\usepackage{graphicx}
\usepackage{amsmath,amssymb} 
\usepackage{color}

\usepackage{microtype}
\usepackage{graphicx}
\usepackage{subcaption}
\usepackage{booktabs} 
\usepackage{makeidx}
\usepackage{amsmath,amssymb} 
\usepackage{color}
\usepackage{listings}

\graphicspath{{figs/}}

\usepackage{algorithm}
\usepackage{algorithmic}
\usepackage{times}
\usepackage{epsfig}
\usepackage{stix}
\usepackage{makecell}
\usepackage{enumitem}

\begin{document}
	\pagestyle{headings}
	\mainmatter

	\title{Deep Archetypal Analysis}

	\titlerunning{Deep Archetypal Analysis}
	\authorrunning{S. M. Keller et al.}
\author{
Sebastian Mathias Keller\inst{1}\orcidID{0000-0002-4531-2839} \thanks{The 16-digit number is the ORCID ID.} \and
Maxim Samarin\inst{1}\orcidID{0000-0002-9242-1827}
Mario Wieser\inst{1}\orcidID{0000-0002-8737-3605} \and
Volker Roth\inst{1}\orcidID{0000-0003-0991-0273}
}

\institute{
$^1$ University of Basel, Switzerland
}

	\maketitle

\input{a00_deepAT_ABSTRACT_gcpr19.tex}
\input{a01_deepAT_INTRO_gcpr19.tex}
\input{a02_deepAT_THEORY_gcpr2019.tex}
\input{a03_deepAT_EXPERIMENTS_Artificial_gcpr2019.tex}

\input{a03_deepAT_EXPERIMENTS_CelebA_gcpr2019.tex}
\input{a03_deepAT_EXPERIMENTS_Molecules_gcpr2019.tex}
\input{a04_deepAT_CONCLUSION_gcpr2019.tex}

\bibliography{bib_cite}
\bibliographystyle{splncs04}

\end{document}

%% file: a00_deepAT_ABSTRACT_gcpr19.tex
\begin{abstract}
Deep Archetypal Analysis (DeepAA) generates latent re\-presentations of high-dimensional datasets in terms of intuitively understandable basic entities called \textit{archetypes}. The proposed method extends linear Archetypal Analysis (AA), an unsupervised method to represent multivariate data points as convex combinations of extremal data points. Unlike the original formulation, Deep AA is generative and capable of handling side information. In addition, our model provides the ability for data-driven representation learning which reduces the dependence on expert knowledge. We empirically demonstrate the applicability of our approach by exploring the chemical space of small organic molecules. In doing so, we employ the archetype constraint to learn two different latent archetype representations for the \textit{same} dataset, with respect to two chemical properties. This type of supervised exploration marks a distinct starting point and let us steer de novo molecular design.








\end{abstract}

%% file: a01_deepAT_INTRO_gcpr19.tex
\section{Introduction}
\label{sec_intro}

Archetypal analysis (AA) is of particular interest when a given data set is assumed to be a superposition of various populations or mechanisms. For a given number of $k$ archetypes, linear AA finds an optimal approximation of the data convex hull, i.e. a polytope, with respect to a given loss function. All data points can then be described as convex mixtures of these $k$ extreme points. In evolutionary biology this has led to the interpretation of archetypes as the representatives most adapted to a given task while non-archetypal representatives are described as mixtures of these extreme or pure types -- able to perform a variety of tasks but non of them optimally \cite{shoval2012}. We identified several limitations of the linear AA model which we would like to address. (I) For data points on a linear submanifold, e.g. a plane in $\mathbb{R}^2$, a strictly monotone transformation should in general have no influence on which points are identified as archetypes. But such a transformation would in fact introduce a non-zero curvature to that submanifold. As a consequence it would become impossible for \textit{linear} AA to approximate equally well the data convex hull, given the same number $k$ of archetypes as before. (II) Using linear AA to explore a dataset and uncover meaningful archetypes usually requires some form of prior knowledge. Either by knowing how many archetypes $k$ are necessary to have an acceptable trade-off between interpretability and error \textit{or} by having domain knowledge about which dimensions of the dataset can/should be omitted, scaled or combined. This procedure of injecting \textit{side information} to the exploration of a given dataset is unpractical at best, impossible even if no intuition about a given problem can be formed. Often side information is available in form of scalar labels, but of course side information could be any kind of richly structured data. When learning a representation of the data, linear AA offers no possibility to incorporate such side information by which to guide the selection of an optimal number of archetypes as well as the relevant dimensions. (III) Linear AA is non-generative. But especially the prospect of incorporating differentiated side information makes the ability to generate new samples -- conditioned on that side inormation -- more attractive. Closely related to that is the ability to interpolate which, in the sense of AA, would be expressed as a \textit{geometric} interpolation within a coordinate system spanned by the $k$ archetypes.\\
In the following we will propose solutions to these limitations in order to extend the area of applicability of AA. In short this entails recasting linear AA as a latent space model within the framework of the Deep Variational Information Bottleneck. Of course this means that our extension constitutes a non-linear version of AA. Extentions into non-linearity have been proposed in the past based on kernalization but such frameworks remain less flexible still in comparison to a learned deep network architecture. But with this increase in flexibility comes a certain trade-off: without side information to guide the learning of a meaningful latent representation the result might be without significance as increased flexibility implies a multitude of possible latent representations dependent only on the side information on which any learning process should therefore be conditioned.\\

\subsubsection{Literature} Linear ``Archetypal Analysis'' (AA) was first proposed by Adele Cutler and Leo Breiman \cite{cutlerBreiman1994}. Since its conception AA has known several advancements on the algorithmic as well as the application side. An extension to (non-linear) Kernel AA is proposed by \cite{bauck2014Kernel,morupKernelAA}, algorithmic improvements by adapting a Frank--Wolfe type algorithm to calculate the archetypes are made by \cite{bauck2015FW} and the extension by \cite{seth2016} introduces a probabilistic version of AA. Archetypal style analysis \cite{styleAA} applies AA to the learned image representations in deep neural networks for artistic style manipulation. In \cite{sand2012} the authors are concerned with model selection by asking for the optimal number of archetypes for a given dataset while \cite{kauf2015} addresses in part the shortcoming of AA we describe in the introduction under (ii). Although AA did not prevail as a commodity tool for pattern analysis it has for example been used by \cite{bauck2009ImgCol} to find archetypal images in large image collections or by \cite{can2015} to perform the analogous task for large document collections. For the human genotype data studied by \cite{huggins2007}, inferred archetypes are interpreted as representative populations for the measured genotypes. And in \cite{chan2003} AA is used to analyse galaxy spectra which are viewed as weighted superpositions of the emissions from stellar populations, nebular emissions and nuclear activity.  
%
%
Our work builds upon Variational Autoencoders (VAEs), arguably the most prevalent representatives of the class of ``Deep Latent Variable Models''. VAEs were introduced by \cite{kingmaWelling2013,rezende2014} and use an inference network to perform a variational approximation of the posterior distribution of the latent variable. Important work in this direction include \cite{KingmaSemi,pmlrrezende15} and \cite{Jang}. More recently, \cite{alemi2016} has discovered a close connection between VAE and the Information Bottleneck principle \cite{tishby2000}. Here, the Deep Variational Information Bottleneck (DVIB) is a VAE where $X$ is replaced by $Y$ in the decoder. Subsequently, the DVIB has been extended in multiple directions such as sparsity \cite{2018arXiv180406216W} or causality \cite{2018arXiv180702326P}.\\
At the same time as our work, \textit{AAnet} was published by \cite{AAnet}. There the authors introduce a neural network based extentension of linear archetypal analysis on the basis of standard non-variational autoencoders. In their work two regularization terms, applied to an intermediate representation provide the latent archetypal convex representation of a non-linear transformation of the input. In contrast to our work which is based on probabilistic generative models (VAE, DVIB), \textit{AAnet} attempts to emulate the generative process by adding noise to the latent representation during training. Further, no side information is incorporated which can -- and in our opinion should -- be used to constrain potentially \textit{over-flexible} neural networks and guide the optimisation process towards learning a meaningful representation. 
%
%
\subsubsection{Contribution}We propose \textit{Deep Archetypal Analysis} (DeepAA) which is a novel, non--linear extension of the original model proposed by \cite{cutlerBreiman1994}. By introducing \textit{DeepAA} within a DVIB framework we address several issues of the original model. Unlike the original model, \textit{DeepAA} (i) is able to identify meaningful archetypes even on non-linear data manifolds, (ii) does not rely on expert knowledge when combining relevant dimensions or learning appropriate transformations (e.g. scaling) and (iii) is able to incorporate side information into the learning process in order to regularize and guide the learning process towards meaningful latent representations. On a large scale experiment we demonstrate the usefulness of \textit{DeepAA} in a setting with side information on the QM9 dataset which contains the chemical structures and properties of 134 kilo molecules \cite{rudd2012,rama2014}. As modern chemistry and material science are increasingly concerned with material property prediction, we show that \textit{DeepAA} can be used to \textit{systematically} explore vast chemical spaces in order to identify starting points for further chemical optimisation.

%% file: a02_deepAT_THEORY_gcpr2019.tex
\section{Method}
\label{sec_theo}
%

\subsection{Linear Archetypal Analysis}
Linear AA \cite{cutlerBreiman1994} is a form of non-negative matrix factorization where a matrix $X\in \mathbb{R}^{n\times p}$ of $n$ data vectors is approximated as $X\approx ABX = AZ$ with $A\in \mathbb{R}^{n\times k}$, $B\in \mathbb{R}^{k\times n}$, and usually $k < \min\{n,p\}$. In AA parlance, the \textit{archetype} matrix $Z\in \mathbb{R}^{k\times p}$ contains the $k$ archetypes $\mathbf{z}_1,..,\mathbf{z}_j,.., \mathbf{z}_k$ and the model is subject to the following constraints:
%
\begin{equation}\label{eq:constraint_A_B}
a_{ij} \geq 0 \text{ }\wedge\text{ } \sum_{j=1}^{k} a_{ij} = 1, \quad 
b_{ji} \geq 0 \text{ }\wedge\text{ } \sum_{i=1}^{n} b_{ji} = 1
\end{equation}
Constraining the entries of $A$ and $B$ to be non-negative and demanding that both weight matrices are row stochastic, implies a representation of the data vectors $\mathbf{x}_{i=1..n}$ as a weighted sum of the rows of $Z$ while simultaneously representing the archetypes $\mathbf{z}_{j=1..k}$ themselves as a weighted sum of the $n$ data vectors in $X$:
\begin{equation} \label{eq:at_decomp}
\mathbf{x}_i \approx \sum_{j=1}^{k} a_{ij} \mathbf{z}_j = \mathbf{a}_i Z, \quad
\mathbf{z}_j = \sum_{i=1}^{n} b_{ji} \mathbf{x}_i  = \mathbf{b}_j X
\end{equation}
Due to the constraints on $A$ and $B$ in Eq. \ref{eq:constraint_A_B} both the representation of $\mathbf{x}_i$ and $\mathbf{z}_j$ in Eq. \ref{eq:at_decomp} are \textit{convex} combinations. Therefore the archetypes approximate the data convex hull and increasing the number $k$ of archetypes improves this approximation. The central problem of AA is finding the weight matrices $A$ and $B$ for a given data matrix $X$.

A probabilistic formulation of linear AA is provided in \cite{seth2016} where it is observed that AA follows a simplex latent variable model and normal observation model. The generative process for the observations $\mathbf{x}_i$ in the presence of $k$ archetypes with archetype weights $\textbf{a}_i$ is given by
\begin{equation}\label{eq:probAT_1}
\mathbf{a}_i \sim \text{Dir}_k(\boldsymbol{\alpha}) \quad \text{ }\wedge\text{ } \quad
\mathbf{x}_i \sim \mathcal{N}(\mathbf{a}_iZ,\,\epsilon^2 \mathbf{I}),
\end{equation}
with uniform concentration parameters $\alpha_j = \alpha$ for all $j$ summing up to $\mathbf{1}^\top\boldsymbol{\alpha}=1$. That is the observations $\mathbf{x}_i$ are distributed according to an isotropic Gaussian with means $\boldsymbol{\mu}_i=\mathbf{a}_iZ$ and variance $\epsilon^2$. 




%

\subsection{Deep Variational Information Bottleneck}
We propose a model to generalise linear AA to the non-linear case based on the Deep Variational Information Bottleneck framework since it allows to incorporate side information $Y$ by design and is known to be equivalent to the VAE in the case of $Y=X$, as shown in \cite{alemi2016}. In contrast to the data matrix $X$ in linear AA, a non-linear transformation $f(X)$ giving rise to a latent representation $T$ of the data suitable for (non-linear) archetypal analysis is considered. I.e. the latent representation $T$ takes the role of the data $X$ in the previous treatment.\\
The DVIB combines the information bottleneck (IB) with the VAE approach \cite{tishby2000,kingmaWelling2013}. The objective of the IB method is to find a random variable $T$ which, while compressing a given random vector $X$, preserves as much information about a second given random vector $Y$. The objective function of the IB is as follows
\begin{equation}
\mbox{min}_{p(\mathbf{t}|\mathbf{x})} I(X;T) - \lambda I(T;Y),
\label{eq:ib1}
\end{equation}
where $\lambda$ is a Lagrange multiplier and $I$ denotes the mutual information. Assuming the IB Markov chain $T-X-Y$ and a parametric form of Eq. \ref{eq:ib1} with parametric conditionals $p_\phi(\mathbf{t}|\mathbf{x})$ and $p_\theta(\mathbf{y}|\mathbf{t})$, Eq. \ref{eq:ib1} is written as
\begin{equation}
\max_{\phi,\theta}  -I_{\phi}(\mathbf{t};\mathbf{x}) + \lambda  I_{\phi,\theta}(\mathbf{t};\mathbf{y}).
\label{eq:ib_parametricForm}
\end{equation}
As derived in \cite{2018arXiv180406216W}, the two terms in Eq. \ref{eq:ib_parametricForm} have the following forms:
\begin{equation}
\label{eq:encoder_parametric}
I_{\phi}(T;X) = D_{KL}\left( p_\phi(\mathbf{t}|\mathbf{x}) p(\mathbf{x}) \| p(\mathbf{t}) p(\mathbf{x})\right) = \mathbb{E}_{p(\mathbf{x})}  D_{KL}\left(p_\phi(\mathbf{t}|\mathbf{x})\| p(\mathbf{t}) \right)
\end{equation}
and
\begin{equation}
\label{eq:decoder_parametric}
\begin{aligned}
I_{\phi,\theta}(T;Y) &=\ D_{KL}\left(\left[\int p(\mathbf{t}|\mathbf{y},\mathbf{x})p(\mathbf{y},\mathbf{x})\, \mathrm{d}\mathbf{x} \right] \| p(\mathbf{t}) p(\mathbf{y})\right)\\
&=\ \mathbb{E}_{p(\mathbf{x},\mathbf{y})} \mathbb{E}_{p_\phi(\mathbf{t}|\mathbf{x})}\log p_\theta(\mathbf{y}|\mathbf{t}) + h(Y).
\end{aligned}
\end{equation}
Here $h(Y)=-\mathbb{E}_{p(\mathbf{y})}\log p(\mathbf{y})$ denotes the entropy of $Y$ in the discrete case or the differential entropy in the continuous case. The models in Eq. \ref{eq:encoder_parametric} and Eq. \ref{eq:decoder_parametric} can be viewed as the encoder and decoder, respectively. Assuming a standard prior of the form $p(\mathbf{t})=\mathcal{N}(\mathbf{t};0,I)$ and a Gaussian distribution for the posterior $p_\phi(\mathbf{t}|\mathbf{x})$, the KL divergence in Eq. \ref{eq:encoder_parametric} becomes a KL divergence between two Gaussian distributions which can be expressed in analytical form as in \cite{kingmaWelling2013}. $I_\phi(T;X)$ can then be estimated on mini-batches of size $m$ as 
\begin{equation}
\label{eq:vae_encoder}
I_\phi(\mathbf{t};\mathbf{x}) \approx \frac1m \sum_i D_{KL}\left(p_\phi(\mathbf{t}|\mathbf{x}_i)\| p(\mathbf{t}) \right).
\end{equation}
As for the decoder, $\mathbb{E}_{p(\mathbf{x},\mathbf{y})} \mathbb{E}_{p_\phi(\mathbf{t}|\mathbf{x})}\log p_\theta(\mathbf{y}|\mathbf{t})$ in Eq. \ref{eq:decoder_parametric} is estimated using the reparametrisation trick proposed by  \cite{kingmaWelling2013,rezende2014}: 
\begin{equation}
\label{eq:vae_decoder}
I_{\phi,\theta}(\mathbf{t};\mathbf{y}) = \mathbb{E}_{p(\mathbf{x},\mathbf{y})} \mathbb{E}_{\boldsymbol{\varepsilon} \sim \mathcal{N}(0,I)} \sum_i  \log  p_{\theta}\left(\mathbf{y}_i|\mathbf{t}_i = \boldsymbol{\mu}_{i}(\mathbf{x}) + diag\left(\boldsymbol{\sigma}_i(\mathbf{x})\right) \boldsymbol{\varepsilon}\right) +\mbox{const.}
\end{equation}
Note that without loss of generality we can assume $Y=(Y',X)$ in Eq. \ref{eq:ib_parametricForm} and with $Y=X$ the original VAE is retrieved. The former will be used in the section \ref{subsec_ChemVAE} experiment where side information $Y'$ is available.
%
\subsection{Deep Archetypal Analysis}
\begin{figure}[h]
\begin{center}
   \includegraphics[scale=.48]{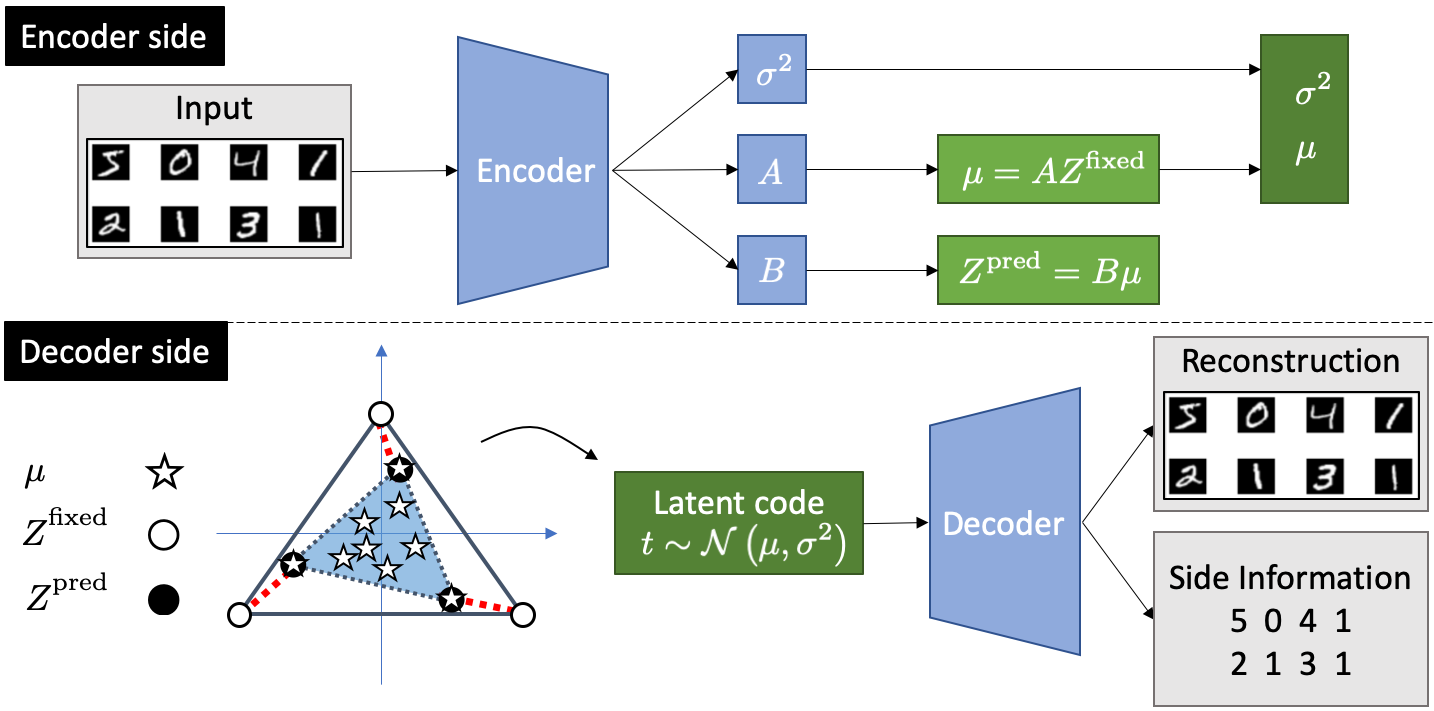}
   
\end{center}
   \caption{Illustration of the \textit{DeepAA} model. \textbf{Encoder side}: Learning weight matrices $A$ and $B$ allows to compute the archetype loss $\ell_{AT}$ in Eq. \eqref{eq:lossAT} and sample latent variables $\mathbf{t}$ as described in Eq. \eqref{eq:deepAA_t}. The constraints on $A$ and $B$ in Eq. \eqref{eq:constraint_A_B} are enforced by using softmax layers. \textbf{Decoder side}: $Z^\text{fixed}$ represent the fixed archetype positions in latent space while $Z^\text{pred}$ are given by the convex hull of the transformed data point means $\mu$ during training. Minimizing $\ell_{AT}$ corresponds to minimizing the red dashed (pairwise) distances. The input is reconstructed from the latent variable $\mathbf{t}$. In the presence of side information, the latent representation allows to reproduce the side information $Y'$ as well as the input $X$.}
\label{fig:at-supervised-arch}
\end{figure}

Deep Archetypal Analysis can then be formulated in the following way. For the sampling of $\mathbf{t}_i$ in Eq. \eqref{eq:vae_decoder} the probabilistic AA approach as in Eq. \eqref{eq:probAT_1} can be used which leads to
\begin{equation}
\label{eq:deepAA_t}
\mathbf{t}_i \sim \mathcal{N}\left(\boldsymbol{\mu}_i(\mathbf{x})=\mathbf{a}_iZ,\,\boldsymbol{\sigma}_i^2(\mathbf{x}) \mathbf{I}\right),
\end{equation}
where the mean $\boldsymbol{\mu}_i$ given through $\mathbf{a}_i$ and variance $\boldsymbol{\sigma}_i^2$ are non-linear transformations of the data point $\mathbf{x}_i$ learned by the encoder. We note that the means $\boldsymbol{\mu}_i$ are convex combinations of weight vectors $\mathbf{a}_i$ and the archetypes $\mathbf{z}_{j=1..k}$ which in return are considered to be convex combinations of the means $\boldsymbol{\mu}_{i=1..m}$ and weight vectors $\mathbf{b}_j$.\footnote{Note that $i=1..m$ (and not up to $n$), which reflects that deep neural networks usually require batch-wise training with batch size $m$.} By learning weight matrices $A\in \mathbb{R}^{m\times k}$ and $B\in \mathbb{R}^{k\times m}$ which are subject to the constraints formulated in Eq. \eqref{eq:constraint_A_B} and parameterised by $\phi$, a non-linear transformation of data $X$ is learned which drives the structure of the latent space to form archetypes whose convex combination yield the transformed data points. A major difference to linear AA is that for \textit{DeepAA} we cannot identify the positions of the archetypes $\mathbf{z}_j$ as there is no absolute frame of reference in latent space. We thus position $k$ archetypes at the vertex points of a $(k-1)$-simplex and collect these \textit{fixed} coordinates in the matrix $Z^{\text{fixed}}$. These requirements lead to an additional archetypal loss of
\begin{equation}\label{eq:lossAT}
\ell_{\text{AT}} = ||Z^{\text{fixed}}-BAZ^{\text{fixed}}||_2^2 = ||Z^{\text{fixed}}-Z^{\text{pred}}||_2^2,
\end{equation}
where $Z^\text{pred} = BAZ^{\text{fixed}}$ are the \textit{predicted} archetype positions given the learned weight matrices $A$ and $B$. For $Z^\text{pred}\approx Z^\text{fixed}$ the loss function $\ell_{\text{AT}}$ is minimized and the desired archetypal structure is achieved. The objective function of \textit{DeepAA} is then given by 
%
%
\begin{equation}
\max_{\phi,\theta}  -I_{\phi}(\mathbf{t};\mathbf{x}) + \lambda  I_{\phi,\theta}(\mathbf{t};\mathbf{y})-\ell_{\text{AT}}.
\label{eq:ib_DeepAA}
\end{equation}
A visual illustration of \textit{DeepAA} is given in Fig. \ref{fig:at-supervised-arch}. The constraints on $A$ and $B$ can be guaranteed by using softmax layers and \textit{DeepAA} can be trained with a standard stochastic gradient descent technique such as Adam \cite{KingmaB14}.

%% file: a03_deepAT_EXPERIMENTS_Artificial_gcpr2019.tex
\section{Experiments}
\label{sec_exp}
\subsection{Artificial Experiments}
\label{subsec_artificial}
\subsubsection{Data generation}
For our experiments we generate data $\mathbf{X}\in \mathbb{R}^{n\times 8}$ that are a convex mixture of $k$ archetypes $\mathbf{Z}\in \mathbb{R}^{k\times 8}$ with $k\ll n$. The generative process for the data $\mathbf{x_i}$ follows Eq. \eqref{eq:probAT_1} where $\mathbf{a}_i$ are stochastic weight vectors denoting the fraction of each of the $k$ archetypes $\mathbf{z}_j$ needed to represent the data point $\mathbf{x}_i$. Here, we generate $n=10000$ data points of which $k=3$ are true archetypes. We set the variance to $\sigma^2=0.05$. We embed our linear 3-dim data manifolfd in a $n=8$ dimensional space. Note that although classical and deep archetypal analysis is always performed on the full data set we only use a fraction of the data when visualizing our results.
\subsubsection{Linear archetypal analysis -- linear data}
\begin{figure}[h!]
	\begin{center}
		\includegraphics[scale=.3]{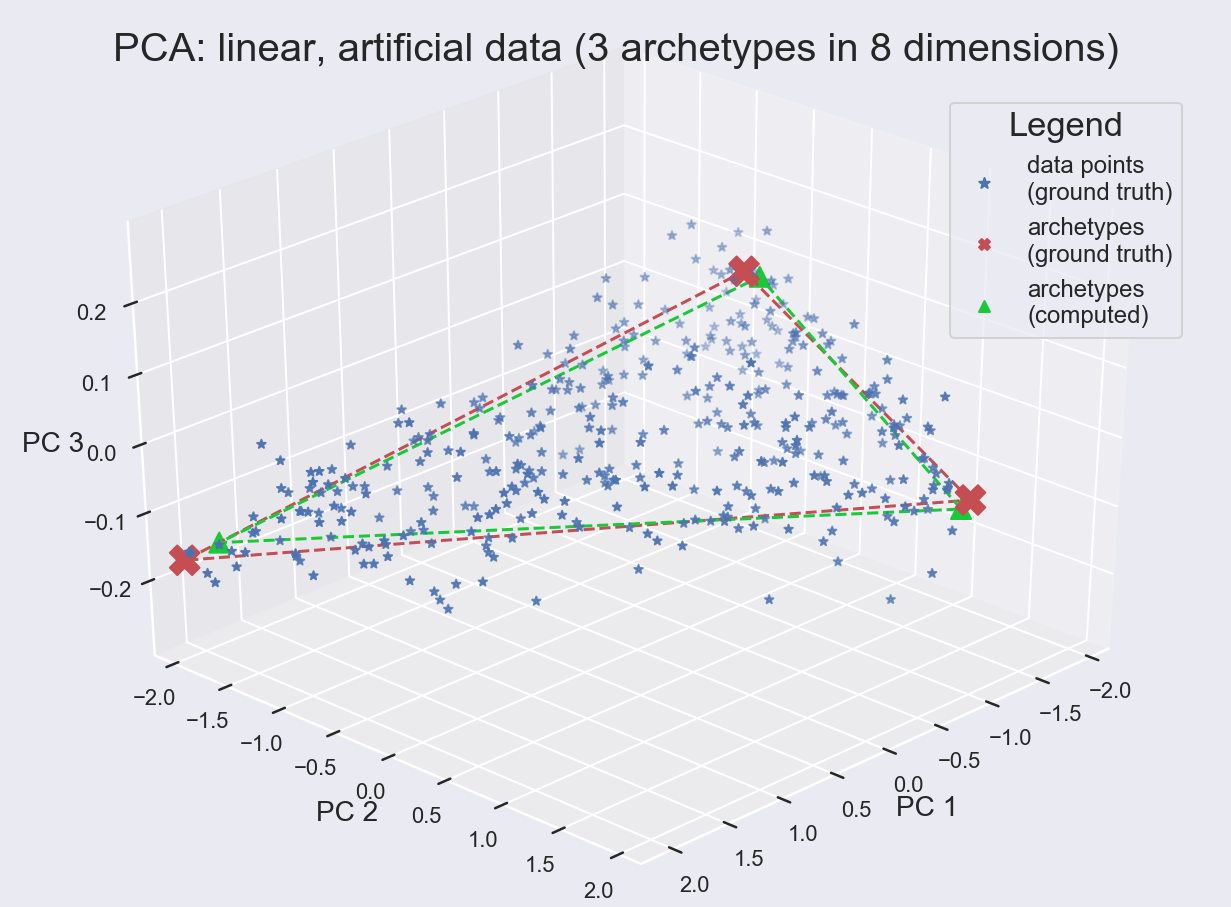}
	\end{center}
	\vspace{-1.8em}
	\caption{PCA projection of 8-dim data after performing linear archetypal analysis. The original linear data submanifold is a convex combination of 3 archetypes.}
	\label{fig:art1}
\end{figure}

Linear archetypal analysis is performed using the efficient Frank-Wolfe procedure proposed in \cite{bauck2015FW}. The input data is 8-dimensional and consequently the dimensionality of the archetypes is $\mathbf{Z}\in \mathbb{R}^{3\times 8}$. For visualization we then use PCA to recover the original 3-dimensional manifold. The first three principal components of the ground truth data are shown in Fig. \ref{fig:art1} as well as the computed archetypes (green triangles). The positions of the computed archetypes are in very good agreement with the ground truth.
In these experiments, the data generating process is known and the number of archetypes $k=3$ can be considered as an available \textit{side information}.
%
%
\subsubsection{Linear archetypal analysis -- non-linear data}
\begin{figure}[h!]
  \begin{subfigure}[t]{.48\textwidth}
  \centering
    \includegraphics[width=.95\textwidth]{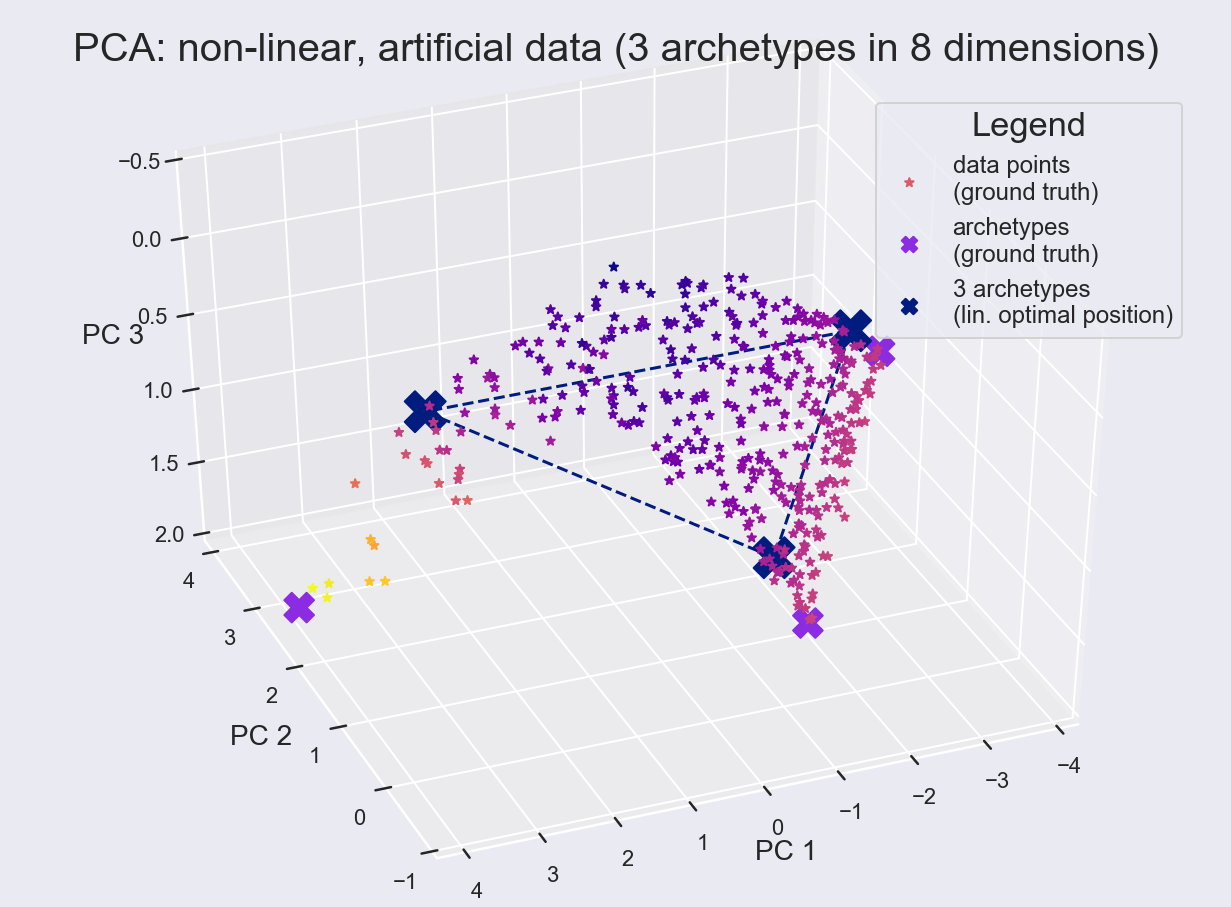}
    \caption{A curved 2-dim manifold. None of the three archetypes identified by linear archetypal analysis can be interpreted as extremes.}
    \label{fig:curvedSubmanifold3ATs}
  \end{subfigure}\hfill
  \begin{subfigure}[t]{.48\textwidth}
  \centering
    \includegraphics[width=.95\textwidth]{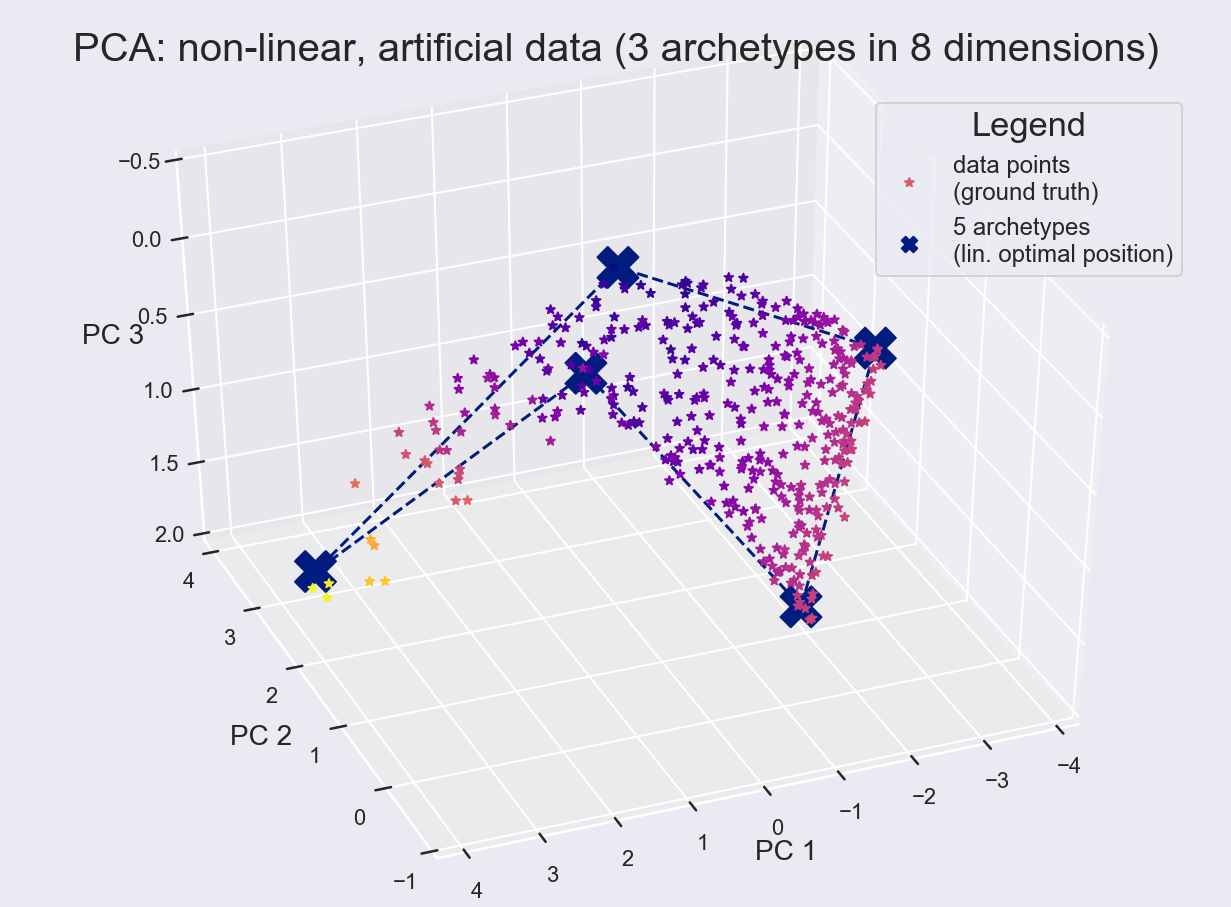}
    \caption{Linear archetypal analysis requires at least five archetypes to describe the data convex hull reasonably well.}
    \label{fig:curvedSubmanifold5ATs}
  \end{subfigure}
  \caption{While linear archetypal analysis is in general able to approximate the data convex hull given a large enough number of archetypes, their interpretation as extremal elements is in general not ensured.}
\end{figure}

Introducing a non-linearity to the data, e.g. by applying the exponential to a dimension of $\mathbf{X}$, results in a curved data submanifold as shown in Fig. \ref{fig:curvedSubmanifold3ATs}.
For example ratios of power or field quantities are usually measured in decibels which is the logarithm of the these ratios. An exponentiation, i.e. introducing a strictly monotone transform, should in general not change which data points are identified as archetypes nor the number of archetypes necessary to obtain a given loss value. Fig. \ref{fig:curvedSubmanifold3ATs} demonstrates that linear archetypal analysis is unable to recover the true archetypes on the same dataset used in the previous experiment but \textit{after} a strictly monotone transform had been applied. Moreover to obtain a similar reconstruction loss as for the \textit{linear} submanifold at least 5 archetypes are necessary as can be seen in Fig. \ref{fig:curvedSubmanifold5ATs}. Although the additional two archetypes are necessary to better aprroximate the data convex hull it would be counter-intuitive to interpret them as \textit{extremes} of the dataset.
\subsubsection{Non-linear archetypal analysis -- non-linear data}
Deep archetypal analysis \textit{without explicit} side information is used to learn a latent linear archetypal representation. We consider as implicit side information the knowledge that 3 archetypes were used to generate our artificial non-linear data and therefore chose a 2-dim latent space. 
\begin{figure}[h!]
  \begin{subfigure}[t]{.48\textwidth}
  \centering
    \includegraphics[width=.95\textwidth]{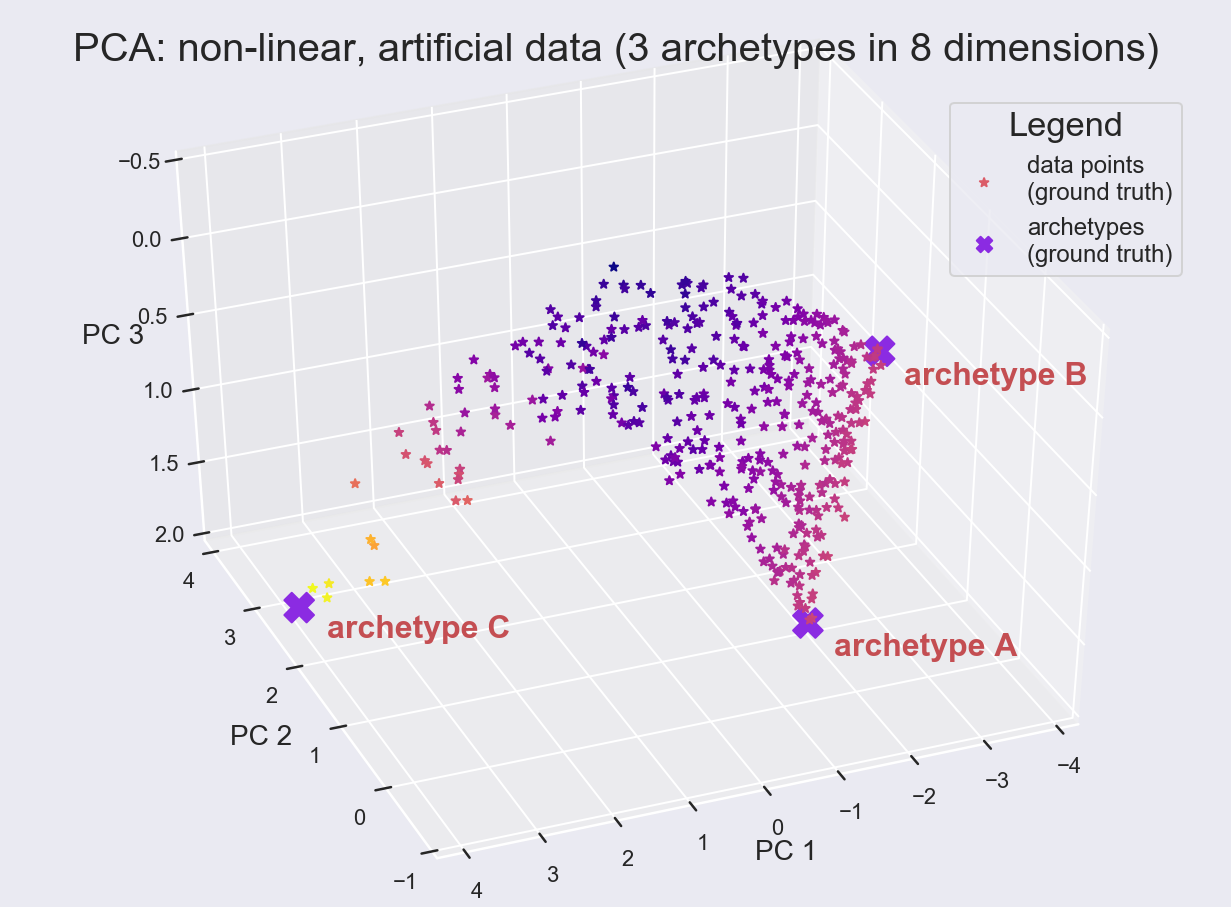}
    \caption{The first three principal components of a non-linear 8-dim manifold. Despite the curvature only three \textit{true} archetypes exist.}
    \label{fig:deep_curvedSubmanifold3ATs}
  \end{subfigure}\hfill
  \begin{subfigure}[t]{.48\textwidth}
  \centering
    \includegraphics[width=1.00\textwidth]{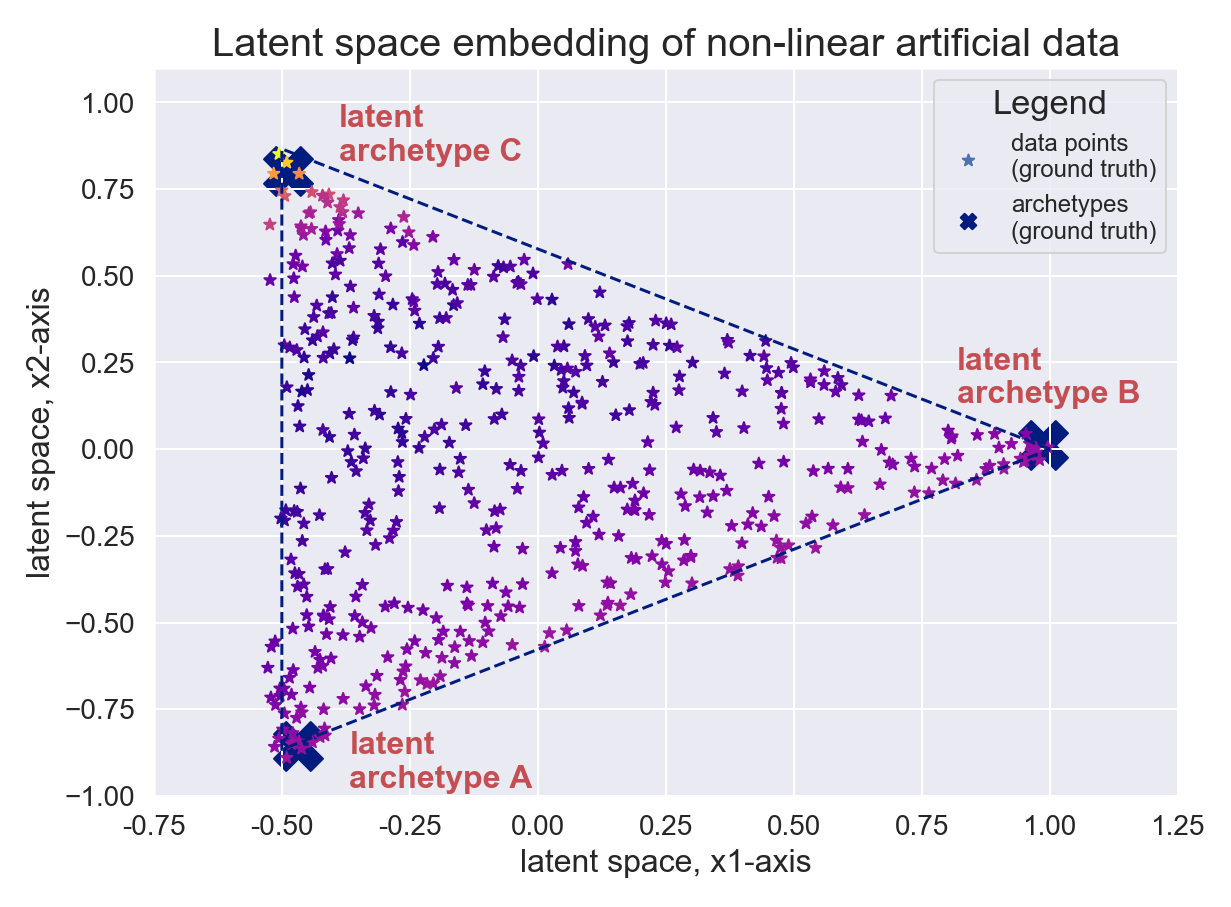}
    \caption{2-dim latent space learned by deep archetypal analysis. Side information used was the known number of archetypes.}
    \label{fig:deepLatentSpce3ATs}
  \end{subfigure}
  \caption{Deep archetypal analysis maps the archetypes from data space onto the vertices of the simplex and conserves the stripe pattern visible in the data space.}
\end{figure}
In Fig. \ref{fig:deep_curvedSubmanifold3ATs} the first three principal components of the 8-dim data are shown. Data points have been colored according to the third principal component. In Fig. \ref{fig:deepLatentSpce3ATs} the learned latent space shows that the archetypes A, B and C have been mapped to the appropriate vertices of the latent simplex. Moreover the sequence of color stripes shown in Fig. \ref{fig:deep_curvedSubmanifold3ATs} has correctly been mapped into latent space. Within the latent space data points are again described as convex linear combinations of the latent archetypes. Latent data points can also be reconstructed in the original data space through the learned decoder network. The network architecture used for this experiment was a simple feedforward network (2 layered encoder and decoder), training for 20 epochs with a batch size of 100 and a learning rate of 0.001.

%% file: a03_deepAT_EXPERIMENTS_CelebA_gcpr2019.tex
\subsection{Generative Aspects and Model Selection}
\label{subsec_celebA}
\textit{DeepAA} allows to generate samples by specifying the mixture coefficients or proportions each archetype shall have in the make-up of a new sample. As a proof of concept, archetypal faces in the large-scale CelebFaces Attributes (CelebA) dataset \cite{celebA2015} are learned and new faces generated.\\
In our experiment we adopt the "Deep Feature Consistent Variational Autoencoder" proposed by \cite{dfc_vae} which makes use of a (feature) perceptual loss as the reconstruction loss.  In our implementation, we use the VAE-123 model of the original paper with the modification as depicted in Fig. \ref{fig:at-supervised-arch}. We train our model with the Adam optimizer \cite{KingmaB14} at a learning rate of $0.0005$ and we set the first moment decay rate to $\beta_1=0.5$. Training is performed with a batch size of 64 for 10 epochs and 90\%/10\% split of the dataset for training/testing. In the experiment, no side information was used. In order to identify the appropriate number of archetypes we propose a model selection technique similar to the ''elbow'' method by \cite{hart2015}: The (minimal) reconstruction loss for different numbers of archetypes, evaluated on the test set, is recorded as shown in Fig. \ref{fig:celebA_IC}. The optimal number of archetypes is considered to be the point where the curve starts converging, which in our case is at 35 archetypes (archetypal faces can be found in the supplement). Fig. \ref{fig:celebA_seq_images} displays an exemplary interpolation of generated faces: Starting at the latent coordinates which represent the face of a young man we move along a straight line in direction of a vertex point of the latent space simplex. While moving along this line we decode, at regular intervals, a total of six latent samples.
\begin{figure}[!h]
	\centering
	\includegraphics[width=0.8\textwidth]{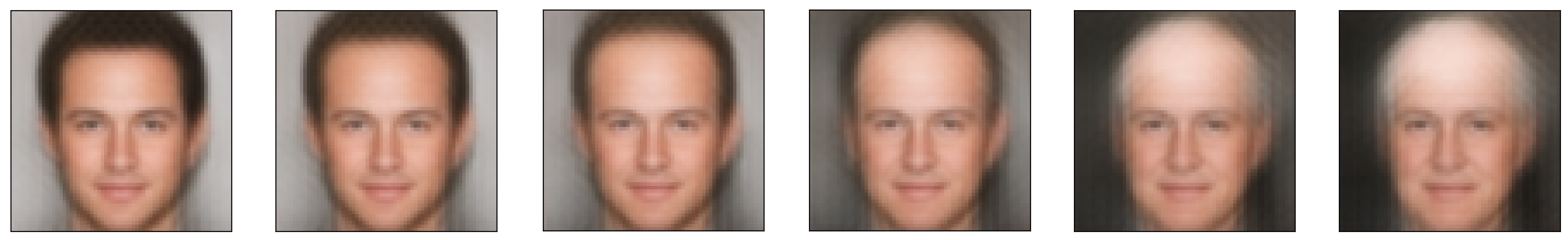}
	\caption{Interpolation sequence towards an archetype representing an old man: While approaching the archetype (archetype B3 in the supplement), characteristic features of the archetypal face are reinforced.}
	\label{fig:celebA_seq_images}
\end{figure}

%% file: a03_deepAT_EXPERIMENTS_Molecules_gcpr2019.tex
%
%
%
%
%
\begin{figure}[hb!]
	\begin{subfigure}[t]{.48\textwidth}
	\centering
		\includegraphics[width=1.0\textwidth]{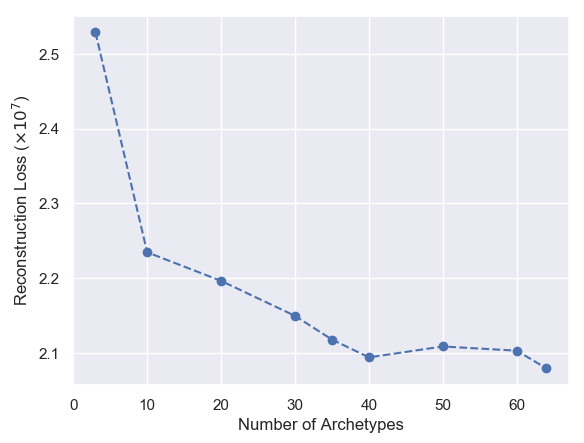}
		\caption{CelebA: Reconstruction loss with varying number of archetypes. } 
		\label{fig:celebA_IC}
	\end{subfigure}\hfill
	\begin{subfigure}[t]{.48\textwidth}
		\centering
		\includegraphics[width=1.0\textwidth]{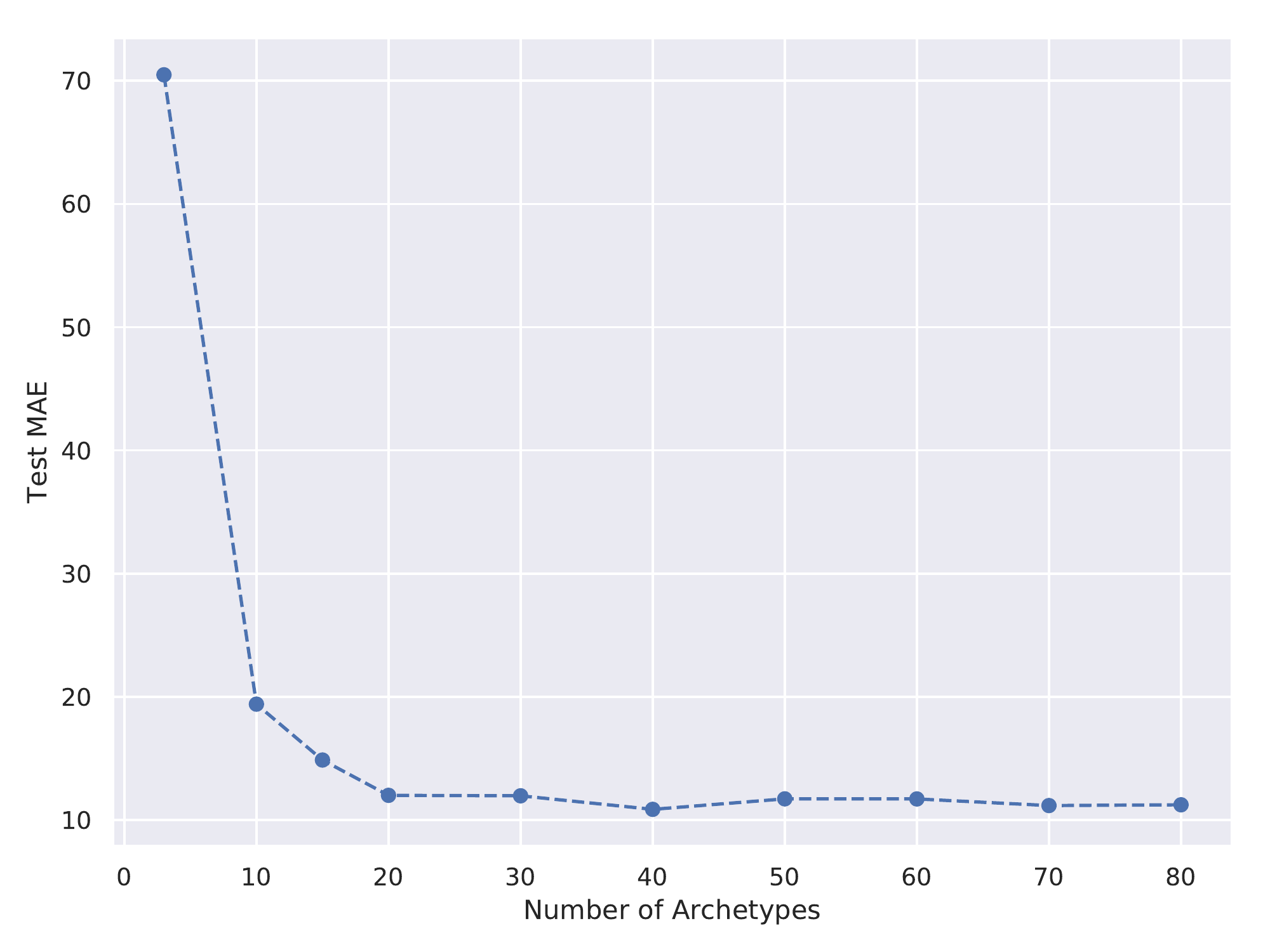}
		\caption{QM9: Test MAE with a varying number of archetypes.} 
		\label{fig:qm9selection}
	\end{subfigure}
\caption{Model selection curves: reconstruction loss vs the number of archetypes.} 
\end{figure}
\begin{figure*}[!h]
    \centering
    \begin{subfigure}{0.45\textwidth}
        \centering
        \includegraphics[width=\textwidth]{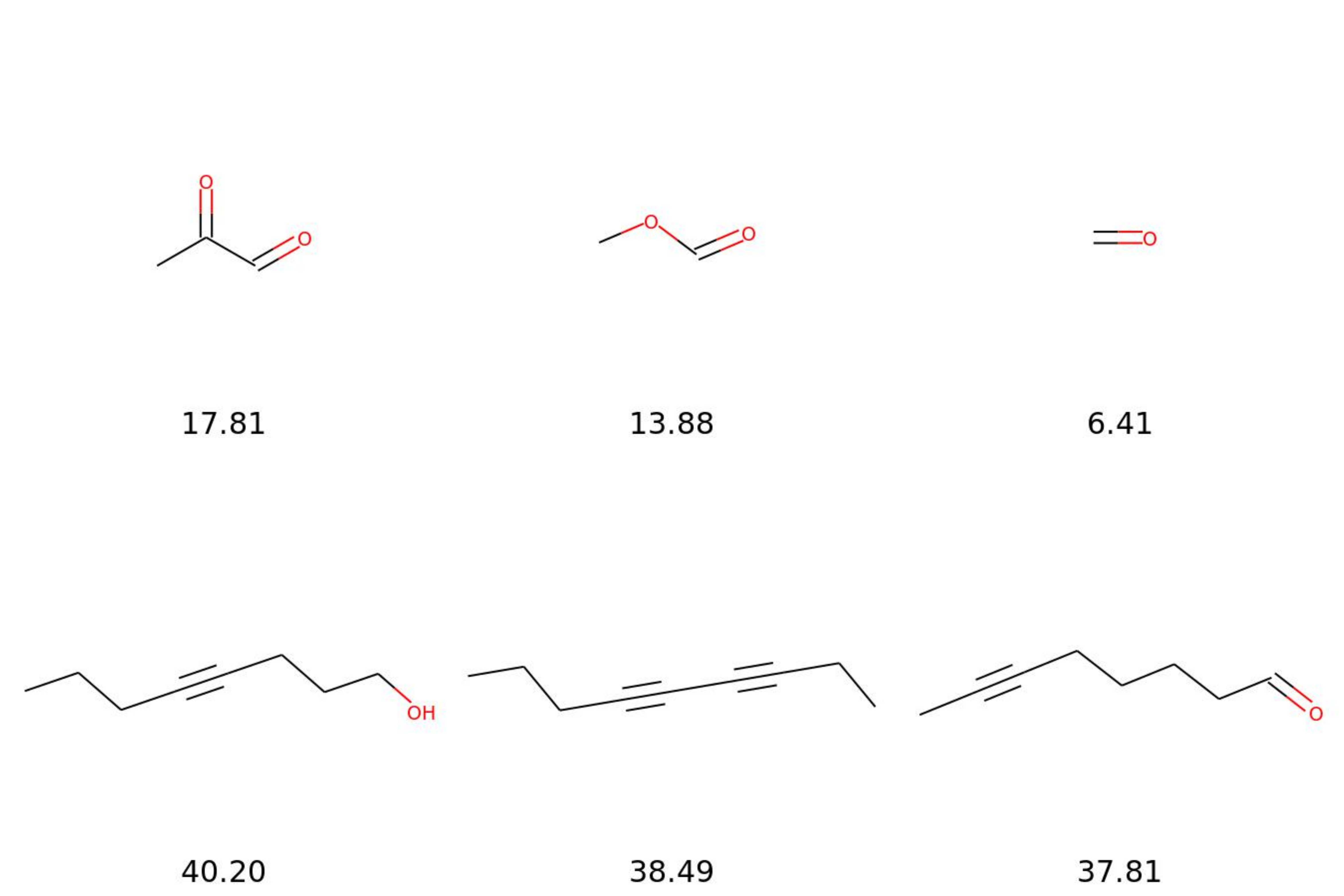}
        \caption{}\label{figure:case1}
    \end{subfigure} \hspace{1cm}
    \begin{subfigure}{0.45\textwidth}
        \centering
        \includegraphics[width=\textwidth]{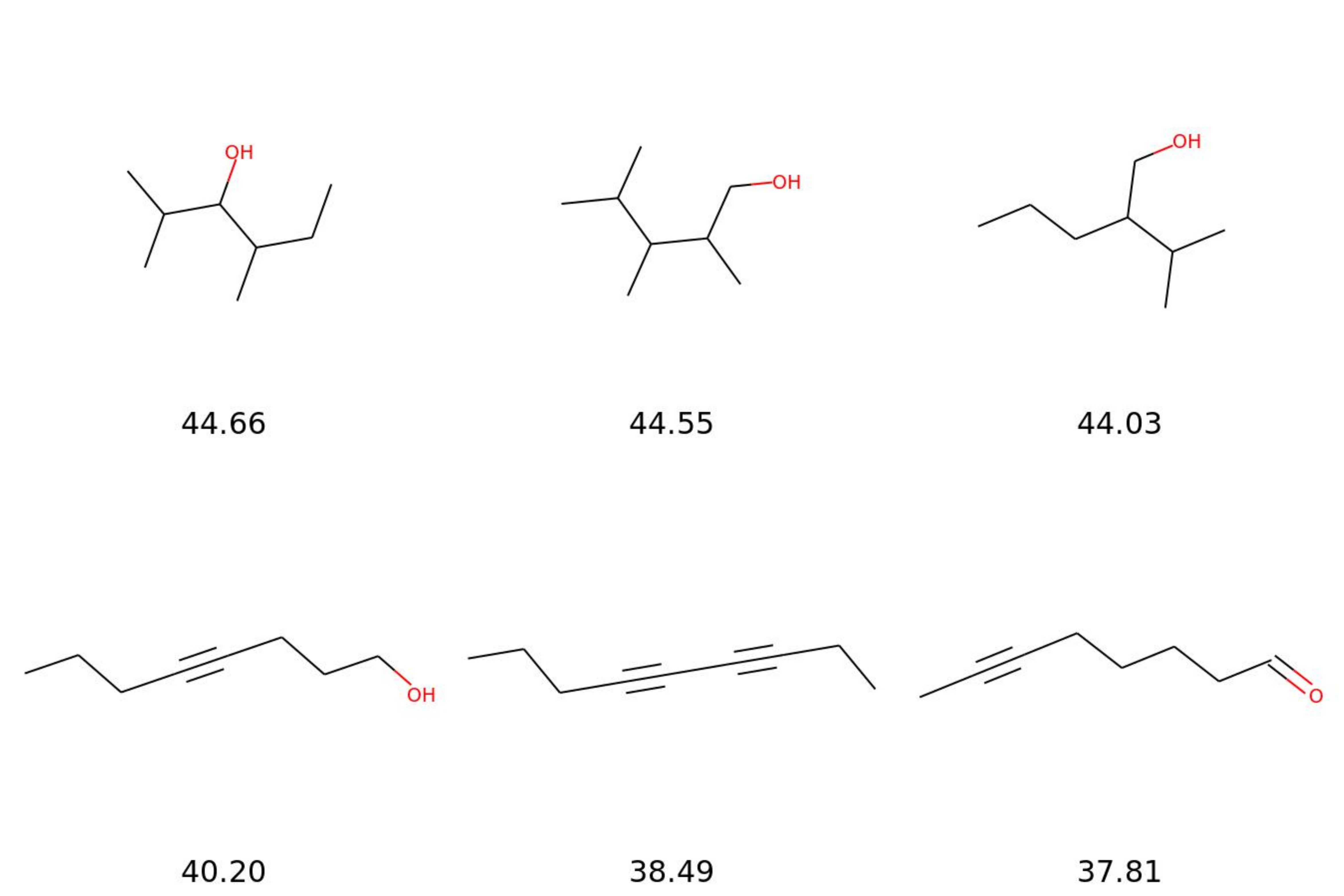}
        \caption{}\label{figure:case2}
    \end{subfigure}
    \caption{The panels illustrate a comparison between two archetypes where the labels represent the corresponding heat capacity. Here, the columns denote the molecules that are closest to the specific archetype and the rows are the archetypes. Panel (a) compares a long chain versus a short chain archetype. Panel (b) compares archetypal molecules with the same mass but different shapes.}
    \label{fig:comparisions}
\end{figure*}

\subsection{Exploring Chemical Spaces with Side Information}
\label{subsec_ChemVAE}
\textbf{Dataset:} As mentioned in the introduction, archetypal analysis lends itself to a distinctly evolutionary interpretation. Although this is certainly a more biological perspective, the basic principle can be transferred to other fields such as chemistry. In this experiment we explore the chemical space which is the space of all molecules that already exist or can be produced. As side information we use the \textit{heat capacity} $C_v$ which quantifies the amount of energy (in Joule) needed to increase 1 Mol of molecules by 1 K at constant volume. Here, a high $C_v$ is especially important for a huge number of applications such as thermal energy storage \cite{CABEZA20151106}. In our experiments, we use the QM9 dataset \cite{rama2014,rudd2012} which was calculated on ab initio DFT method based structures and properties of 134k organic molecules with up to nine atoms (C, O, N, or F), without counting hydrogen.
%
%
%
\begin{figure*}[!h]
\begin{center}
\includegraphics[width=0.9\textwidth]{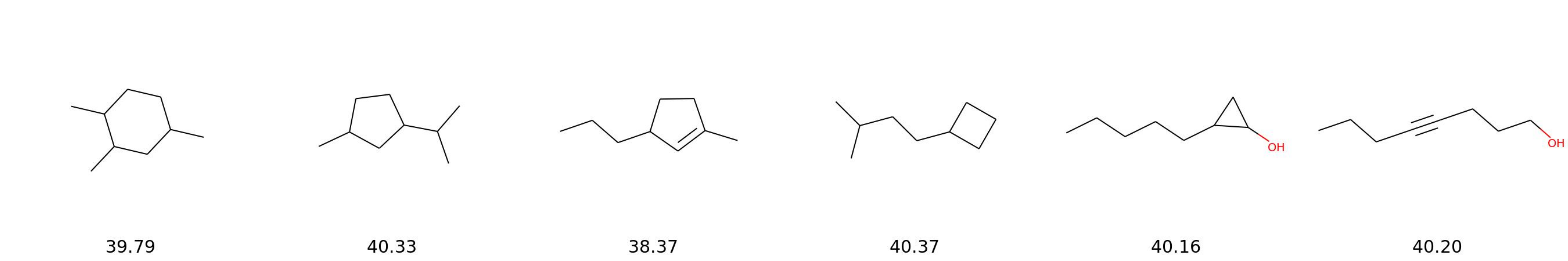}
\caption{Interpolation between two archetypes produced by our model. The label denote the molecules' heat capacity. While we show only one example, the same results can also be observed for other archetype combinations.} 
\label{fig:interpolation}
\end{center}
\end{figure*}

\textbf{Set-up:} We extracted 204 features for every molecule by using the Chemistry Development Kit \cite{cdk}. The neural architectures used have 3 hidden layers with 1024, 512 and 256 neurons, respectively and ReLU activation functions. We train our model in a \textit{supervised} fashion, by reconstructing the molecule and the side information simultaneously. In Experiment 1, we continuously increase the number of latent dimensions to perform model selection. In Experiment 2 and 3, we fix the number of latent dimensions to 19 which corresponds to 20 archetypes. During training, we steadily increase the Lagrange multiplier $\lambda$ by 1.01 every 500 iterations. Our model is trained with the Adam optimizer \cite{KingmaB14} with an initial learning rate of 0.01. We decay the learning rate with an exponential decay by 0.95 every 10k iterations. In addition, we use a batch size of 2048 and train the model for 350k iterations. The dataset is divided in a training and test split of 90/10\%.

In \textbf{Experiment 1}, we asses the MAE error when varying the number of archetypes in Fig. \ref{fig:qm9selection}. In our case, we perform model selection by observing where the MAE converges (starting from 20 archetypes) to select the optimal number of archetypes. Obviously, if the number of archetypes is smaller, it becomes more difficult to reconstruct the data. This stems from the fact there exist a large number of molecules with almost the same heat capacity but with a different shape. Thus, molecules with different shapes are mapped to archetypes with the same heat capacity which makes it hard to resolve the many-to-one mapping in the latent space. 

In \textbf{Experiment 2}, we identify archetypal molecules that are associated with a particular heat capacity. In this setting, we focus on 20 archetypes (Fig. \ref{fig:qm9selection}) to obtain the optimal exploration-exploitation trade-off. While focusing only on a small selection of archetypes, we provide the full list in the supplement. In chemistry, the heat capacity is defined as $C_v = \dfrac{d\epsilon}{dT} \bigm|_{v=const}$ where $\epsilon$ denotes the energy of a molecule and $T$ is the temperature. The energy can be further decomposed into $\epsilon = \epsilon^{Tr}+\epsilon^R+\epsilon^V+\epsilon^E$ where $Tr$ depicts translation, $R$ rotation, $V$ vibration and $E$ the electric contribution, respectively \cite{atkins2010atkins,tinoco2002physical}. Building upon this knowledge, we compare different archetypal molecules associated with a particular heat capacity (Fig. \ref{fig:comparisions}). Here, the rows correspond to archetypes and the columns depict the three closest test molecules to the archetype. In Fig. \ref{figure:case1} we illustrate two archetypes with a high and low heat capacity. The first row archetype has a lower heat capacity because of its shorter chain and more double bonds. Due to these properties, the archetype is more stable which results in a lower vibrational energy $V$ and subsequently in a lower heat capacity. Fig. \ref{figure:case2} plots both a non-linear and a linear archetypal molecule with the same atomic mass. Here, the linear molecule loses one of its rotational modes due to its geometry. Therefore, the second row archetype has a lower rotational energy $R$ compared to the first row archetype, leading to a lower heat capacity. 

In \textbf{Experiment 3}, we focus on the interpolation between two archetypes. We do so by plotting the test samples which are closest to the linear connection between the two archetypes. Here, we observe a smooth transition from a ring molecule to a linear molecule with the same heat capacity. Along these archetypes, which both are similar in heat capacities but differ in shape, a molecule can only change its shape but it cannot go beyond a particular heat capacity. Results are shown in Fig. \ref{fig:interpolation}. 

\begin{figure*}[!h]
    \centering
    \begin{subfigure}{0.45\textwidth}
        \centering
        \includegraphics[width=\textwidth]{figs/long_chain.pdf}
        \caption{}\label{figure:heatcap}
    \end{subfigure} \hspace{.35cm}
    \begin{subfigure}{0.45\textwidth}
        \centering
        \includegraphics[width=\textwidth]{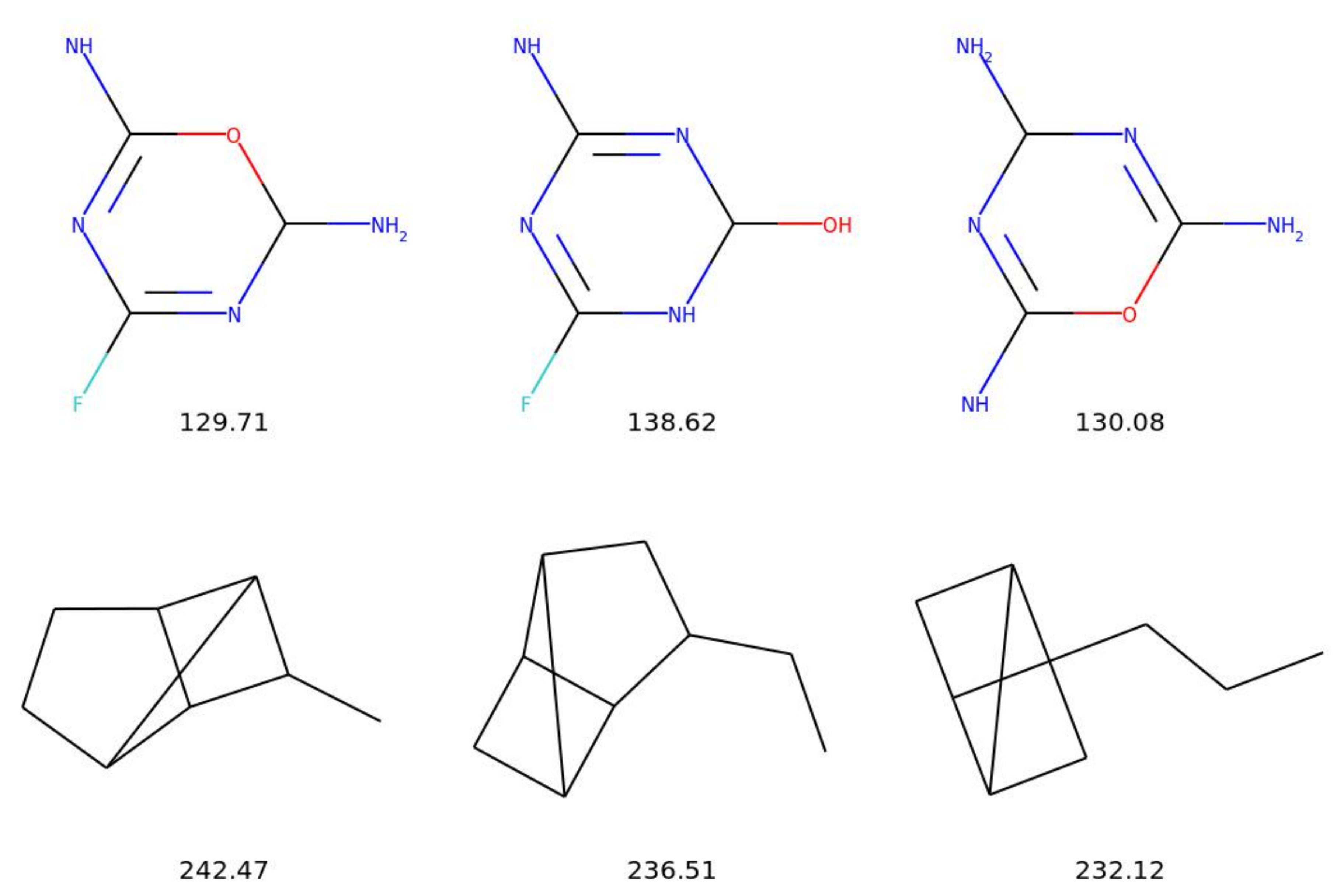}
        \caption{}\label{figure:gap}
    \end{subfigure}
    \caption{The panels illustrate a comparison between archetypes with side information with the highest and lowest property values. Here, the labels correspond to the heat capacity (a) and the band gap energy (b). The columns denote the molecules that are closest to a specific certain and the rows denote the archetypes. Panel (a) depicts archetypal heat capicity molecules and Panel (b) shows archetypal band gap energy molecules.}
    \label{fig:side}
\end{figure*}

Finally, in \textbf{Experiment 4}, we demonstrate that our model structures latent spaces according to the side information provided. Consequently, a molecule being a mxiture of archetypes with respect to heat capacity might become archetypal with respect to another property. Here, we compare the discovered archetypes for two specific chemical properties, heat capacity and band gap energy. In Fig. \ref{fig:side}, we plot the archetypes with the highest and lowest heat capacity (Fig. \ref{figure:heatcap}) and the highest and lowest band gap energy (Fig. \ref{figure:gap}), respectively. The extreme archetypes significantly differ in their structure as well as their atomic composition based on their property. For example, the archetype with low heat capacity are rather small with only a few C and O atoms. In contrast, the archetype with a low band gap energy are composed as rings with N and H atoms. A more detailed comparison between all archetypes can be found in the supplement. 

%% file: a04_deepAT_CONCLUSION_gcpr2019.tex
\section{Conclusion}
\label{sec_concl}
In this paper, we introduced a novel neural network based approach to learn a structured latent representation of a given dataset. The structure we impose onto the latent space allows to characterize this space through its most extremal or \textit{archetypal} representatives. In doing so, we build upon the linear AA approach and combine this concept with the deep IB principle to obtain a non-linear archetype model. In contrast to the classical approach our method offers three advantages: First, our model introduces a data-driven representation learning which reduces the dependence on expert knowledge. Second, we learn appropriate transformations to obtain meaningful archetypes even on non-linear data manifolds. Third, we are able to incorporate side information into the learning process. This counteracts overly flexible deep neural networks in order to identify meaningful archetypes with specific properties and facilitate an interpretable exploration of the latent space representations. Our experiment on the QM9 molecular dataset demonstrate the applicability of our method in an important real world setting. 


\paragraph{\textbf{Acknowledgements.}}
S.K. is partially supported by the the Swiss National Science Foundation project CR32I2 159682. M.S. is supported by the Swiss National Science Foundation grant 407540 167333 as part of the Swiss National Research Programme NRP 75 "Big Data". M.W. is partially supported by the NCCR MARVEL, funded by the Swiss National Science Foundation and SNSF grant 51MRP0158328 (SystemsX.ch).

%% file: 128-main.bbl
\begin{thebibliography}{10}
\providecommand{\url}[1]{\texttt{#1}}
\providecommand{\urlprefix}{URL }
\providecommand{\doi}[1]{https://doi.org/#1}

\bibitem{alemi2016}
Alemi, A.A., Fischer, I., Dillon, J.V., Murphy, K.: Deep variational
  information bottleneck. CoRR  \textbf{abs/1612.00410} (2016),
  \url{http://arxiv.org/abs/1612.00410}

\bibitem{atkins2010atkins}
Atkins, P., de~Paula, J.: Atkins' Physical Chemistry. OUP Oxford (2010)

\bibitem{bauck2014Kernel}
Bauckhage, C., Manshaei, K.: Kernel archetypal analysis for clustering web
  search frequency time series. In: 2014 22nd International Conference on
  Pattern Recognition. pp. 1544--1549 (Aug 2014). \doi{10.1109/ICPR.2014.274}

\bibitem{bauck2015FW}
Bauckhage, C., Kersting, K., Hoppe, F., Thurau, C.: Archetypal analysis as an
  autoencoder. In: Workshop New Challenges in Neural Computation 2015. pp.
  8--16 (10 2015),
  \url{https://www.techfak.uni-bielefeld.de/~fschleif/mlr/mlr\_03\_2015.pdf}

\bibitem{bauck2009ImgCol}
Bauckhage, C., Thurau, C.: Making archetypal analysis practical. In: Denzler,
  J., Notni, G., S{\"u}{\ss}e, H. (eds.) Pattern Recognition. pp. 272--281.
  Springer Berlin Heidelberg (2009)

\bibitem{CABEZA20151106}
Cabeza, L.F., Gutierrez, A., Barreneche, C., Ushak, S., Fernandez, A.G.,
  Fernadez, A.I., Grageda, M.: Lithium in thermal energy storage: A
  state-of-the-art review. Renewable and Sustainable Energy Reviews
  \textbf{42},  1106 -- 1112 (2015)

\bibitem{can2015}
Canhasi, E., Kononenko, I.: Weighted hierarchical archetypal analysis for
  multi-document summarization. Computer Speech \& Language  \textbf{37} (11
  2015). \doi{10.1016/j.csl.2015.11.004}

\bibitem{cutlerBreiman1994}
Cutler, A., Breiman, L.: Archetypal analysis. Technometrics  \textbf{36}(4),
  338--347 (1994). \doi{10.1080/00401706.1994.10485840},
  \url{http://digitalassets.lib.berkeley.edu/sdtr/ucb/text/379.pdf}

\bibitem{AAnet}
van Dijk, D., Burkhardt, D., Amodio, M., Tong, A., Wolf, G., Krishnaswamy, S.:
  Finding archetypal spaces for data using neural networks. arXiv preprint
  arXiv:1901.09078  (2019)

\bibitem{chan2003}
H.~P.~Chan, B., Mitchell, D., Cram, L.: Archetypal analysis of galaxy spectra.
  Monthly Notices of the Royal Astronomical Society  \textbf{338} (01 2003).
  \doi{10.1046/j.1365-8711.2003.06099.x}

\bibitem{hart2015}
Hart, Y., Sheftel, H., Hausser, J., Szekely, P., Ben-Moshe, N.B., Korem, Y.,
  Tendler, A., Mayo, A.E., Alon, U.: Inferring biological tasks using pareto
  analysis of high-dimensional data. Nature methods  \textbf{12}(3), ~233
  (2015)

\bibitem{dfc_vae}
Hou, X., Shen, L., Sun, K., Qiu, G.: Deep feature consistent variational
  autoencoder. In: Applications of Computer Vision (WACV), 2017 IEEE Winter
  Conference on. pp. 1133--1141. IEEE (2017)

\bibitem{huggins2007}
Huggins, P., Pachter, L., Sturmfels, B.: Toward the human genotope. Bulletin of
  Mathematical Biology  \textbf{69}(8),  2723--2735 (Nov 2007).
  \doi{10.1007/s11538-007-9244-7},
  \url{https://doi.org/10.1007/s11538-007-9244-7}

\bibitem{Jang}
{Jang}, E., {Gu}, S., {Poole}, B.: {Categorical Reparameterization with
  Gumbel-Softmax}. International Conference on Learning Representations (ICLR)
  (2017)

\bibitem{kauf2015}
Kaufmann, D., Keller, S., Roth, V.: Copula archetypal analysis. In: Gall, J.,
  Gehler, P., Leibe, B. (eds.) Pattern Recognition. pp. 117--128. Springer
  International Publishing (2015)

\bibitem{KingmaB14}
Kingma, D.P., Ba, J.: Adam: A method for stochastic optimization.
  \textbf{abs/1412.6980} (2014)

\bibitem{KingmaSemi}
Kingma, D.P., Mohamed, S., Rezende, D.J., Welling, M.: Semi-supervised learning
  with deep generative models. In: Advances in Neural Information Processing
  Systems 27: Annual Conference on Neural Information Processing Systems 2014,
  December 8-13 2014, Montreal, Quebec, Canada. pp. 3581--3589 (2014)

\bibitem{kingmaWelling2013}
Kingma, D.P., Welling, M.: Auto-encoding variational bayes. CoRR
  \textbf{abs/1312.6114} (2013)

\bibitem{celebA2015}
Liu, Z., Luo, P., Wang, X., Tang, X.: Deep learning face attributes in the
  wild. In: Proceedings of International Conference on Computer Vision (ICCV)
  (December 2015)

\bibitem{morupKernelAA}
M{\o}rup, M., Hansen, L.K.: Archetypal analysis for machine learning and data
  mining. Neurocomputing  \textbf{80},  54--63 (2012)

\bibitem{2018arXiv180702326P}
{Parbhoo}, S., {Wieser}, M., {Roth}, V.: {Causal Deep Information Bottleneck}.
  arXiv e-prints arXiv:1807.02326 (Jul 2018)

\bibitem{sand2012}
Prabhakaran, S., Raman, S., Vogt, J.E., Roth, V.: Automatic model selection in
  archetype analysis. In: Pinz, A., Pock, T., Bischof, H., Leberl, F. (eds.)
  Pattern Recognition. pp. 458--467. Springer Berlin Heidelberg (2012)

\bibitem{rama2014}
Ramakrishnan, R., Dral, P.O., Rupp, M., von Lilienfeld, O.A.: Quantum chemistry
  structures and properties of 134 kilo molecules. Scientific Data  \textbf{1}
  (2014)

\bibitem{pmlrrezende15}
Rezende, D., Mohamed, S.: Variational inference with normalizing flows. In:
  Bach, F., Blei, D. (eds.) Proceedings of the 32nd International Conference on
  Machine Learning. Proceedings of Machine Learning Research, vol.~37, pp.
  1530--1538. PMLR, Lille, France (07--09 Jul 2015)

\bibitem{rezende2014}
Rezende, D.J., Mohamed, S., Wierstra, D.: Stochastic backpropagation and
  approximate inference in deep generative models  \textbf{32}(2),  1278--1286
  (22--24 Jun 2014)

\bibitem{rudd2012}
Ruddigkeit, L., van Deursen, R., Blum, L.C., Reymond, J.L.: Enumeration of 166
  billion organic small molecules in the chemical universe database gdb-17.
  Journal of Chemical Information and Modeling  \textbf{52}(11),  2864--2875
  (2012). \doi{10.1021/ci300415d},
  \url{https://pubs.acs.org/doi/10.1021/ci300415d}, pMID: 23088335

\bibitem{seth2016}
Seth, S., Eugster, M.J.A.: Probabilistic archetypal analysis. Machine Learning
  \textbf{102}(1),  85--113 (Jan 2016). \doi{10.1007/s10994-015-5498-8},
  \url{https://doi.org/10.1007/s10994-015-5498-8}

\bibitem{shoval2012}
Shoval, O., Sheftel, H., Shinar, G., Hart, Y., Ramote, O., Mayo, A., Dekel, E.,
  Kavanagh, K., Alon, U.: Evolutionary trade-offs, pareto optimality, and the
  geometry of phenotype space. Science  \textbf{336}(6085),  1157--1160 (2012).
  \doi{10.1126/science.1217405},
  \url{http://science.sciencemag.org/content/336/6085/1157}

\bibitem{cdk}
Steinbeck, C., Han, Y.Q., Kuhn, S., Horlacher, O., Luttmann, E., Willighagen,
  E.: {The Chemistry Development Kit (CDK): An open-source Java library for
  chemo- and bioinformatics}. {Journal of Chemical Information and Computer
  Sciences}  \textbf{43}(2),  493--500 (2003)

\bibitem{tinoco2002physical}
Tinoco, I.: Physical Chemistry: Principles and Applications in Biological
  Sciences. No. S. 229-313 in Physical Chemistry: Principles and Applications
  in Biological Sciences, Prentice Hall (2002)

\bibitem{tishby2000}
Tishby, N., Pereira, F.C., Bialek, W.: The information bottleneck method. arXiv
  preprint physics/0004057  (2000)

\bibitem{2018arXiv180406216W}
{Wieczorek}, A., {Wieser}, M., {Murezzan}, D., {Roth}, V.: {Learning Sparse
  Latent Representations with the Deep Copula Information Bottleneck}.
  International Conference on Learning Representations (ICLR)  (2018)

\bibitem{styleAA}
Wynen, D., Schmid, C., Mairal, J.: Unsupervised learning of artistic styles
  with archetypal style analysis. In: Advances in Neural Information Processing
  Systems. pp. 6584--6593 (2018)

\end{thebibliography}
